%% file: entity-free-uschema.tex
\documentclass{article} % For LaTeX2e
\usepackage{naaclhlt2016}
\usepackage{times}
\usepackage{latexsym}
\usepackage{hyperref}
\usepackage{url}
\usepackage{amsmath,amsthm,amsfonts}
\usepackage{multirow}
\usepackage{xspace}
\usepackage{tikz}
\usetikzlibrary{shapes,backgrounds,patterns}
\usepackage{graphicx}
\usepackage{caption}
\usepackage{subcaption}

\newcommand{\citep}[1]{\cite{#1}}
\newcommand{\citet}[1]{\newcite{#1}}

\title{Row-less Universal Schema}

\author{Patrick Verga \& Andrew McCallum \\
    College of Information and Computer Sciences\\
    University of Massachusetts Amherst\\
    \texttt{\{pat, mccallum\}@cs.umass.edu} \\
}

\begin{document}

\maketitle

\begin{abstract}
Universal schema jointly embeds knowledge bases and textual patterns to reason about entities and relations for automatic knowledge base construction and information extraction.
In the past, entity pairs and relations were represented as learned vectors with compatibility determined by a scoring function, limiting generalization to unseen text patterns and entities.
Recently, `column-less' versions of Universal Schema have used compositional pattern encoders to generalize to all text patterns.
In this work we take the next step and propose a `row-less' model of universal schema, removing explicit entity pair representations.
Instead of learning vector representations for each entity pair in our training set, we treat an entity pair as a function of its relation types.
In experimental results on the FB15k-237 benchmark we demonstrate that we can match the performance of a comparable model with explicit entity pair representations using a model of attention over relation types.
We further demonstrate that the model performs with nearly the same accuracy on entity pairs never seen during training.
\end{abstract}

% intro + background
\input{intro}

% methods
\input{model}

% results
\input {results}

% conclusion
\section{Conclusion}
In this paper we explore a row-less extension of Universal Schema that forgoes explicit entity pair representations for an aggregation function over relation types.
This extension allows prediction between all entity pairs in new textual mentions -- whether seen at train time or not -- and also provides a natural connection to the provenance supporting the prediction.

In this work we show that an aggregation function based on query-specific attention over relation types outperforms query independent aggregations.
We show that aggregation models are able to predict on par with entity pair models for seen entity pairs and, in the case of attention, suffer very little loss for unseen entity pairs.

We also limited our pattern encoders to lookup-tables.
In future work we will combine the column-less and row-less approach to make a fully compositional Universal Schema model.
This will allow Universal Schema to generalize to all new textual patterns and entity pairs.

\subsubsection*{Acknowledgments}
We thank Emma Strubell, David Belanger, Luke Vilnis, and Arvind Neelakantan for helpful discussions and edits.
This work was supported in part by the Center for Intelligent Information Retrieval and the Center for Data Science, in part by The Allen Institute for Artificial Intelligence, and in part by DARPA under agreement number FA8750-13-2-0020. The U.S. Government is authorized to reproduce and distribute reprints for Governmental purposes notwithstanding any copyright notation thereon, in part by Defense Advanced Research Agency (DARPA) contract number HR0011-15-2-0036, and in part by the National Science Foundation (NSF) grant number IIS-1514053. Any opinions, findings and conclusions or recommendations expressed in this material are those of the authors and do not necessarily reflect those of the sponsor.

\bibliography{sources}
\bibliographystyle{naaclhlt2016}

\newpage
%\appendix
%\input{appendix}

\end{document}

%% file: intro.tex
\section{Introduction\label{introduction}}

Automatic knowledge base construction (AKBC) is the task of building a structured knowledge base (KB) of facts using raw text evidence, and often an initial seed KB to be augmented~\citep{NELL,yago,freebase}.
Extracted facts about entities and their relations are useful for many downstream tasks such as question answering and query understanding.
An effective approach to AKBC is Universal Schema, in which relation extraction is modeled as a matrix factorization problem wherein each row of the matrix is an entity pair and each column represents a relation between entities. Relations derived from a KB schema and from free text are thus embedded into a shared space allowing for a rich representation of KB relations, the union of all KB schemata.

This formulation is still limited in terms of its generalization, however.
In its original form, Universal Schema can reason only about entity pairs and text relations explicitly seen at train time; it cannot predict relations between new entity pairs.
In this work we present a `row-less' extension of Universal Schema. Rather than representing each entity pair with an explicit dense vector, we encode entity pairs as aggregate functions over their relation types.
This allows Universal Schema to form predictions for all entity pairs regardless of whether that pair was seen during training, and provides a direct connection between the prediction and its provenance.
% this is breaking compilation
%We demonstrate that our relation only models perform competitively with models using explicit entity pair vectors.

Many models exist which address this issue by operating at the level of entities rather than entity pairs. A knowledge base is naturally described as a graph, in which entities are nodes and relations are labeled edges~\citep{yago,freebase}.
In the case of \emph{knowledge graph completion}, the task is akin to link prediction, assuming an initial set of (\emph{s, r, o}) triples.
See~\citet{nickel2015review} for a review.
No accompanying text data is necessary, since links can be predicted using properties of the graph, such as transitivity.
In order to generalize well, prediction is often posed as low-rank matrix or tensor factorization.
A variety of model variants have been suggested, where the probability of a given edge existing depends on a multi-linear form~\citep{rescal,DBLP:journals/corr/Garcia-DuranBUG15,bishan,transe,wang2014knowledge,lin2015learning}, or non-linear interactions between $s$, $r$, and $o$~\citep{socherkb}. 
These entity-based models have recall advantages over entity pairs.
The model can predict relations between any two entity pairs in the absence of  other information such as the pair's contextual occurrence in text.

However, entity models have been shown to be less precise than entity pair models when text is used to augment knowledge base facts. 
 Both~\citet{toutanova2015representing} and~\citet{limin} observe that the entity pair model outperforms entity models in cases where the entity pair was seen at training time. 
 Since Universal Schema leverages large amounts of unlabeled text we desire the benefits of entity pair modeling, and row-less Universal Schema facilitates learning entity pairs without the drawbacks of the traditional one-embedding-per-pair approach.
 
In this paper we investigate Universal Schema models without explicit entity pair representations.
Instead, entity pairs are represented using an aggregation function over their relation types.
This allows our model to naturally make predictions about any entity pair in new textual mentions, regardless of whether they were seen at train time additionally giving the model direct access to provenance.
We show that an attention-based aggregation function outperforms several simpler functions and matches a model using explicit entity pairs.
We then demonstrate that these `row-less' models accurately predict on entity pairs unseen during training.

%Universal Schema models entity pairs as an explicit vector.
%We instead treat each entity pair as an aggregate function over each of its relation types.
%This allows us to trivially extend to unseen entity pairs, have a direct link to provenance, and allocate a variable number of parameters per entity pair.

\section {Background}

\subsection {Universal Schema}
\begin{figure}[h]
\centering
\includegraphics[scale=.33]{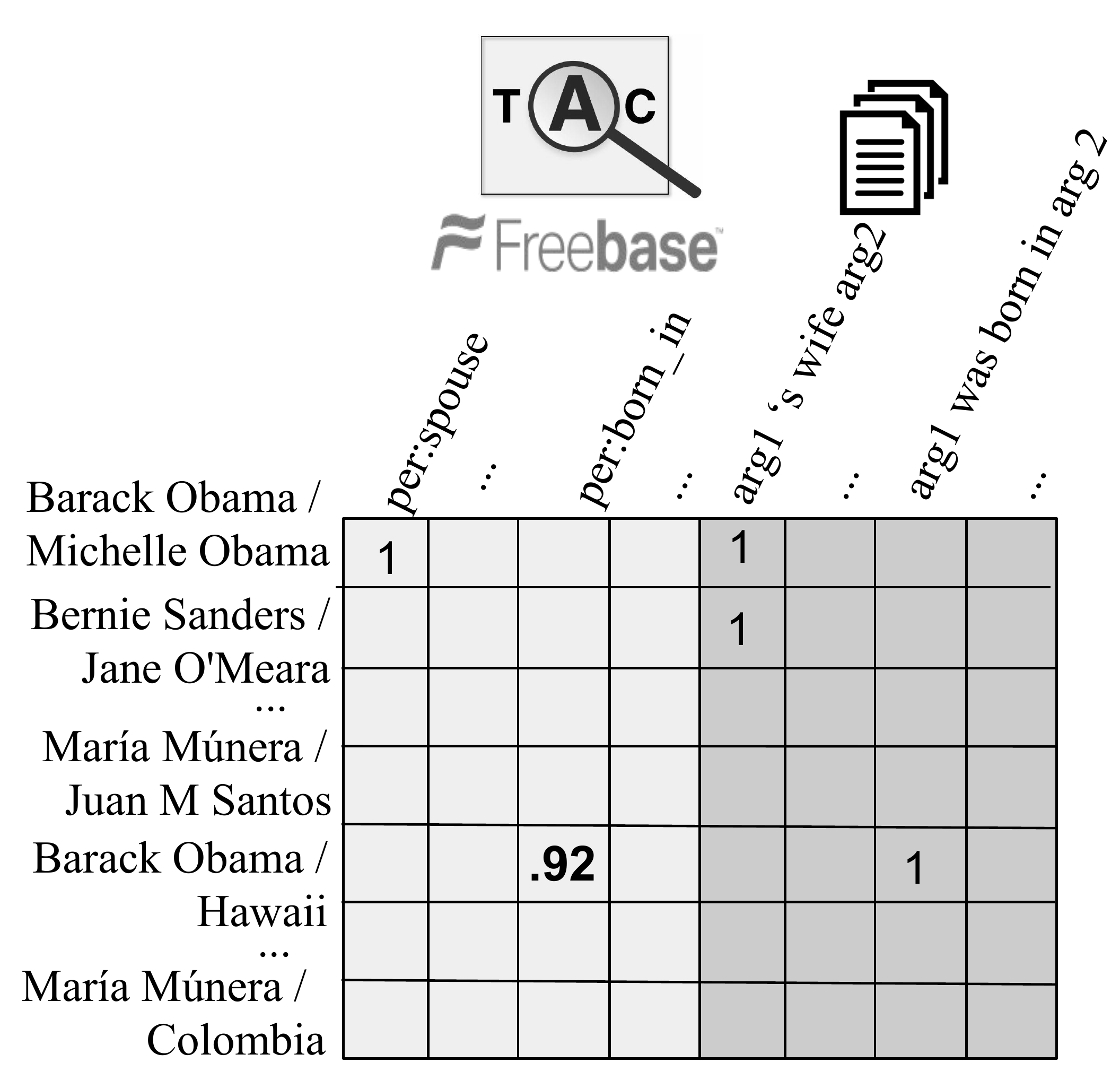}
\caption{Universal schema represents relation types and entity pairs as a matrix.
1s denote observed training examples and the bolded .92 is predicted by the model.
\label{fig:uschema-matrix}}
\end{figure}

The Universal Schema \citep{limin} approach to AKBC jointly embeds any number of KB and text corpora into a shared space to jointly reason over entities and their relations (Figure \ref {fig:uschema-matrix}).
The problem of relation extraction is posed as a matrix completion task where rows are entity pairs and columns are KB relations and textual patterns.
The matrix is decomposed into two low-rank matrices resulting in embeddings for each entity pair, relation, and textual pattern.
Reasoning is then performed directly on these embeddings.

\citet{limin} proposed several model variants operating on entities and entity pairs, and subsequently many other extensions have been proposed \citep{yao2013universal,vector_pra,neelakantan2015compositional,logicmfnaacl15}. Recently, Universal Schema has been extended to encode compositional representations of textual relations \citep{toutanova2015representing,verga2015multilingual} allowing it to generalize to all textual patterns and reason over arbitrary text.

\subsection {`Column-less' Universal Schema}

The original Universal Schema approach has two main drawbacks: similar patterns do not share statistics, and the model is unable to make predictions about textual patterns not explicitly seen at train time.

Recently, `column-less' versions of Universal Schema have been proposed to address these issues \citep{toutanova2015representing,verga2015multilingual}.
These models learn compositional pattern encoders to parameterize the column matrix in place of directly embedding textual patterns. Compositional Universal Schema facilitates more compact sharing of statistics by composing similar patterns from the same sequence of word embeddings -- the text patterns `lives in the city' and `lives in the city of' no longer exist as distinct atomic units. More importantly, Compositional Universal Schema can thus generalize to all possible textual patterns, facilitating reasoning over arbitrary text at test time.

%\begin{figure}[h]
%\vspace{.3cm}
%    \begin{subfigure}[b]{0.3\textwidth}
%    \hspace{1.4cm}
%	\includegraphics[scale=.9]{non-compositional-pattern}
%	\end{subfigure}
%
%    \begin{subfigure}[b]{0.3\textwidth}
%    \vspace{-.5cm}    \hspace{.4cm}
%	\includegraphics[scale=.9]{compositional-pattern}
%	\end{subfigure}
%	\caption{Top: Universal Schema expresses each textual pattern as an atomic unit \protect\citet{limin}.
%Bottom: Compositional Universal Schema uses an LSTM to encode each textual relations \protect\cite{verga2015multilingual}. }
%\end{figure}

%% file: model.tex
%\section{Training a Sentence Classifier without Alignment \label{sec:uschema}}
\section{Model \label{sec:model}}

\subsection {`Row-less' Universal Schema}

While column-less Universal Schema addresses reasoning over arbitrary textual patterns, it is still limited to reasoning over entity pairs seen at training time.
\citet{verga2015multilingual} approach this problem by using Universal Schema as a sentence classifier -- directly comparing a textual relation to a KB relation to perform relation extraction.
However, this approach is unsatisfactory for two reasons.
The first is that this creates an inconsistency between training and testing, as the model is trained to predict compatibility between entity pairs and relations and not relations directly.
Second, it considers only a single piece of evidence in making its prediction.

\begin{figure}[h]
\centering
\includegraphics[scale=.68]{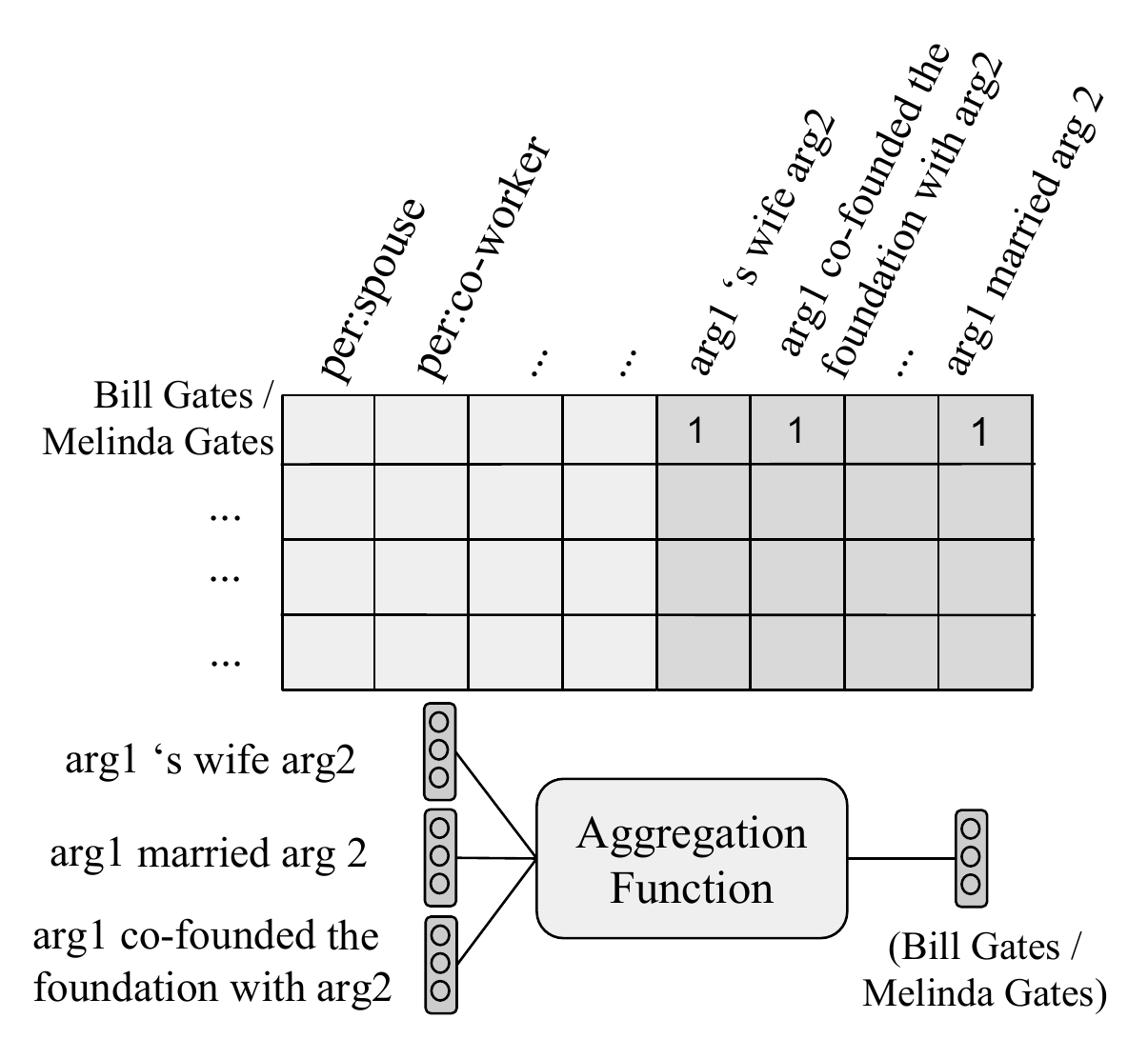}
\caption{Row-less Universal Schema encodes an entity pair as an aggregation of its observed relation types.
\label{fig:aggregation}}
\end{figure}

We address both of these concerns in our `row-less' Universal Schema.
Rather than encoding each entity pair explicitly, we take the compositional approach of encoding entity pairs as an aggregation over their observed relation types (Figure \ref{fig:aggregation}).
A learned entity pair embedding can be seen as a summarization of all relation types for which that entity pair was seen. Rather than learn this summarization as a single embedding, we reconstruct an entity pair representation from an aggregate of its relation types, essentially learning a mixture model rather than a single centroid.
%\todo{describe that its uscham + aggregation - refer to figures}

\subsection {Aggregation Functions \label{sec:functions}}
In this work we examine several  aggregation functions.
\textbf{Mean Pool} creates a single centroid for the entity pair by averaging all of its relation vectors.
While this intuitively makes sense as an approximation for the explicit entity pair representation, averaging large numbers of embeddings can lead to a noisy signal.
\textbf{Max Pool} also creates a single centroid for the entity pair by taking a dimension-wise max over the observed relation type vectors.
Both mean pool and max pool are query-independent and form the same representation for the entity pair regardless of the query relation.

We also examine two query-specific aggregation functions.
These models are more expressive than a single vector that is forced to to act as a centroid to all possible relation types an entity pair can take on.
For example, the entity pair Bill and Melinda Gates could hold the relation `per:spouse' or `per:co-worker'.
A query-specific aggregation mechanism can produce separate representations for this entity pair dependent on the query.

The \textbf{Max Relation} aggregation function represents the entity pair as its most similar relation to the query vector of interest.
This model has the advantage of creating a query-specific entity pair representation but is more susceptible to noisy training data as a single incorrect piece of evidence could be used to form a prediction.
%\textbf {TopK Relations} is a middle ground between Mean and Max relations.

\begin{figure*}[t!]
\hspace{1cm}
\includegraphics[scale=1]{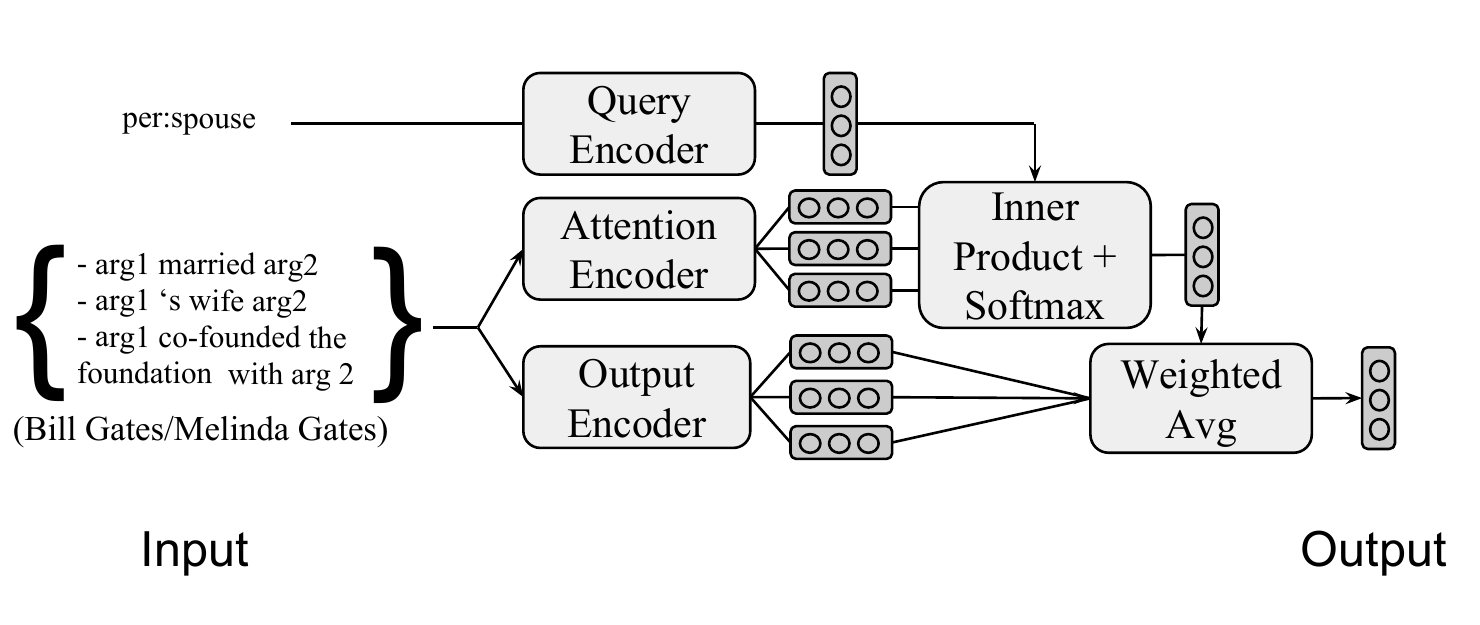}
\caption{\label{fig:attention}
In the Attention model the query is dotted with an input representation of each relation type followed by a softmax, giving a weighting over each relation type.
This output is then used to get a weighted sum over a set of output representations for each relation type.
The result is a query-specific vector representation of the entity pair.
The Max Relation model simply takes the max dot product rather than a softmax and weighted average.
}
\end{figure*}

Finally, we look at an \textbf{Attention} aggregation function over relation types (Figure \ref{fig:attention}) which is similar to a single-layer memory network \cite{sukhbaatar2015end} .
In this model the query is scored with an input representation of each relation type followed by a softmax, giving a weighting over each relation type.
This output is then used to get a weighted sum over a set of output representations for each relation type resulting in a query-specific vector representation of the entity pair.
The model pools relevant information over the entire set of relations and selects the most salient aspects to the query relation.

\subsection {Training}
\citet{limin} use Bayesian Personalized Ranking (BPR)~\citep{rendle2009bpr} to train their Universal Schema models. 
BPR  ranks the probability of observed triples above unobserved triples rather than explicitly modeling unobserved edges as negative.
Each training example is an entity pair/relation type triple observed in the training text corpora or KB.
Rather than BPR, \citet{toutanova2015representing} use a sampled softmax criterion where they use 200 negative samples \footnote{Many past papers restrict negative samples to be of the same type as the positive example.
We simply sample uniformly from the entire set of entities that would form a valid entity pair.} to approximate the negative log likelihood.
Results shown were obtained using sampled softmax which outperformed BPR in our early experiments.

Our training procedure for relation-only Universal Schema model is similar to the original model.
We first pool all of the observed triples in our training data creating entity pair-specific relation sets $R^{Ep}$.
We remove entity pairs with only a single observed relation type.
We then construct training examples for each observed relation type of an entity and an aggregation of all other relation types observed with that entity; for each observed relation type $r_i \in R^{Ep}$, we construct a positive training example $(r_i, \left\{R^{Ep}\setminus r_i\right\})$.
We randomly sample a different relation to act as the negative sample.

All models were implemented in Torch\footnote{\url{https://github.com/patverga/torch-relation-extraction}} and were trained using Adam \cite{kingma2014adam}.
They each were trained with embedding dimension 25 and used 200 negative samples except for max pool which performed better with two negative samples.
The entity pair model used a batch size 1024, $\ell_2 = 1e$-$8$, $\epsilon = 1e$-$4$, and learning rate .01.
The aggregation models all used batch size 4096,  $\ell_2 = 0$, $\epsilon = 1e$-$8$, and learning rate .01.
The column vectors were initialized with the columns learned by the entity pair model.
Randomly initializing the query encoders and tying the output and attention encoders performed better and all results use this method.
Models were tuned to maximize mean reciprocal rank (MRR) on the validation set with early stopping.

%% file: results.tex
\section{Experimental Results\label{sec:results}}

\subsection{Data and Evaluation}

We evaluate our models on the FB15k-237 dataset from \citet{toutanova2015representing}.
The data is composed of a small set of 237 Freebase relations and approximately 4 million textual patterns from Clueweb with entities linked to Freebase \cite{gabrilovich2013facc1}.
In past studies, for each (subject, relation, object) test triple, negative examples are generated by replacing the object with all other entities, filtering out triples that are positive in the data set.
The positive triple is then be ranked among the negatives.
In our experiments we limit the possible generated negatives to those entity pairs that have textual mentions in our training set.
This way we can evaluate how well the model classifies textual mentions as Freebase relations.
We also filter textual patterns with length greater than 35.
We report the percentage of positive triples ranked in the top 10 amongst their negatives as well as the MRR scaled by 100.

\subsection {Results}
Our results are shown in Table \ref{table:seen_results}.
The model with explicit entity pair representations outperforms the `row-less' variant by 2.9 MRR and 2.7\% Hits@10.
However, these results do show that relation aggregation is competitive with entity pair embeddings and it is possible to have a Universal Schema model without entity embeddings.
With further experimentation we are confident that the `row-less' model will perform on-par or better than the entity pair model.

The max model performs competitively with the attention model.
This is not entirely surprising as it is a simplified version of the attention model. 
Further, the attention model reduces to the max relation model for entity pairs with only a single observed relation type.
In our data, 64.8\% of entity pairs have only a single observed relation type and 80.9\% have 1 or 2 observed relation types.

We also explore the models' abilities to predict on unseen entity pairs (Table \ref{table:unseen_results}).
We remove all training examples that contain a positive entity pair in either our validation or test set.
We use the same validation and test set as in Table \ref{table:seen_results}.
The entity pair model predicts random relations as it is unable to make predictions on unseen entity pairs.
The max pool and mean pool each suffer approximately 30\% relative decrease in MRR and 20\% decrease in Hits@10.

Both the max relation and attention models perform nearly as well whether the entity pairs were observed during training or not.
The attention model performs slightly better than the max relation model in this scenario.

\begin{table}[h!]
\setlength{\tabcolsep}{4.1pt}
\begin{center}
\begin{tabular}{|l|cc|}
\hline
\bf Model & MRR & Hits@10 \\
\hline\hline
Entity-pair Embeddings          & 31.85 & 51.72 \\
\hline
Mean Pool                       & 25.89 & 45.94 \\
Max Pool                        & 29.61 & 49.93 \\
Max Relation                    & 31.71 & \bf 51.94 \\
%Attention (seperate)            & 28.83 & 49.46 \\
Attention                       & \bf 31.92 & 51.67 \\
\hline
\end{tabular}
\caption{The percentage of positive triples ranked in the top 10 amongst their negatives as well as the mean reciprocal rank (MRR) scaled by 100 on a subset of the FB15K-237 dataset.
Negative examples were restricted to entity pairs that occurred in the KB or text portion of the training set.
\label{table:seen_results}}
\end{center}
\vspace{-.3cm}
\end{table}

\begin{table}[h!]
\setlength{\tabcolsep}{3pt}
\begin{center}
\begin{tabular}{|l|cc|cc|}
\hline
\bf Model & MRR & \%dec & Hits@10 & \%dec\\
\hline\hline
Entity-pair Embeddings          & 5.23 & 83.6    & 11.94 & 76.9 \\
\hline
Mean Pool                       & 18.10 & 30.1   & 35.76 & 22.2 \\
Max Pool                        & 20.80 & 29.8   & 40.25 & 19.4 \\
Max Relation                    & 28.46 & 10.3   & 48.15 & 7.3 \\
%Attention (seperate)            &  & & & \\
Attention                       & \bf 29.75 & \bf6.8   & \bf 49.69 & \bf3.8 \\
\hline
\end{tabular}
\caption{Predicting entity pairs that were not seen at train time.
The percentage of positive triples ranked in the top 10 amongst their negatives as well as the mean reciprocal rank (MRR) scaled by 100 on a subset of the FB15K-237 dataset. Also shown is the percent relative decrease in MRR and Hits@10 between Table \ref{table:seen_results} (entity pairs seen during training) and this table (entity pairs unseen during training).
\label{table:unseen_results}}
\end{center}
\vspace{-.3cm}
\end{table}

%% file: entity-free-uschema.bbl
\begin{thebibliography}{}

\bibitem[\protect\citename{Bollacker \bgroup et al.\egroup }2008]{freebase}
Kurt Bollacker, Colin Evans, Praveen Paritosh, Tim Sturge, and Jamie Taylor.
\newblock 2008.
\newblock Freebase: a collaboratively created graph database for structuring
  human knowledge.
\newblock In {\em Proceedings of the ACM SIGMOD International Conference on
  Management of Data}.

\bibitem[\protect\citename{Bordes \bgroup et al.\egroup }2013]{transe}
Antoine Bordes, Nicolas Usunier, Alberto Garc{\'\i}a-Dur{\'a}n, Jason Weston,
  and Oksana Yakhnenko.
\newblock 2013.
\newblock Translating embeddings for modeling multi-relational data.
\newblock In {\em Advances in Neural Information Processing Systems.}

\bibitem[\protect\citename{Carlson \bgroup et al.\egroup }2010]{NELL}
Andrew Carlson, Justin Betteridge, Bryan Kisiel, Burr Settles, Estevam~R.
  Hruschka, and A.
\newblock 2010.
\newblock Toward an architecture for never-ending language learning.
\newblock In {\em In AAAI}.

\bibitem[\protect\citename{Gabrilovich \bgroup et al.\egroup
  }2013]{gabrilovich2013facc1}
Evgeniy Gabrilovich, Michael Ringgaard, and Amarnag Subramanya.
\newblock 2013.
\newblock Facc1: Freebase annotation of clueweb corpora, version 1 (release
  date 2013-06-26, format version 1, correction level 0).
\newblock {\em Note: http://lemurproject. org/clueweb09/FACC1/Cited by}, 5.

\bibitem[\protect\citename{Garc{\'{\i}}a{-}Dur{\'{a}}n \bgroup et al.\egroup
  }2015]{DBLP:journals/corr/Garcia-DuranBUG15}
Alberto Garc{\'{\i}}a{-}Dur{\'{a}}n, Antoine Bordes, Nicolas Usunier, and Yves
  Grandvalet.
\newblock 2015.
\newblock Combining two and three-way embeddings models for link prediction in
  knowledge bases.
\newblock {\em CoRR}, abs/1506.00999.

\bibitem[\protect\citename{Gardner \bgroup et al.\egroup }2014]{vector_pra}
Matt Gardner, Partha Talukdar, Jayant Krishnamurthy, and Tom Mitchell.
\newblock 2014.
\newblock Incorporating vector space similarity in random walk inference over
  knowledge bases.
\newblock In {\em Empirical Methods in Natural Language Processing}.

\bibitem[\protect\citename{Kingma and Ba}2015]{kingma2014adam}
Diederik Kingma and Jimmy Ba.
\newblock 2015.
\newblock {Adam: A method for stochastic optimization}.
\newblock In {\em 3rd International Conference for Learning Representations
  (ICLR)}.

\bibitem[\protect\citename{Lin \bgroup et al.\egroup }2015]{lin2015learning}
Yankai Lin, Zhiyuan Liu, Maosong Sun, Yang Liu, and Xuan Zhu.
\newblock 2015.
\newblock Learning entity and relation embeddings for knowledge graph
  completion.
\newblock In {\em Proceedings of AAAI}.

\bibitem[\protect\citename{Neelakantan \bgroup et al.\egroup
  }2015]{neelakantan2015compositional}
Arvind Neelakantan, Benjamin Roth, and Andrew McCallum.
\newblock 2015.
\newblock Compositional vector space models for knowledge base completion.
\newblock {\em Proceedings of the 53rd Annual Meeting of the Association for
  Computational Linguistics}.

\bibitem[\protect\citename{Nickel \bgroup et al.\egroup }2011]{rescal}
Maximilian Nickel, Volker Tresp, and Hans-Peter Kriegel.
\newblock 2011.
\newblock A three-way model for collective learning on multi-relational data.
\newblock In {\em International Conference on Machine Learning.}

\bibitem[\protect\citename{Nickel \bgroup et al.\egroup
  }2015]{nickel2015review}
Maximilian Nickel, Kevin Murphy, Volker Tresp, and Evgeniy Gabrilovich.
\newblock 2015.
\newblock A review of relational machine learning for knowledge graphs: From
  multi-relational link prediction to automated knowledge graph construction.
\newblock {\em arXiv preprint arXiv:1503.00759}.

\bibitem[\protect\citename{Rendle \bgroup et al.\egroup }2009]{rendle2009bpr}
Steffen Rendle, Christoph Freudenthaler, Zeno Gantner, and Lars Schmidt-Thieme.
\newblock 2009.
\newblock Bpr: Bayesian personalized ranking from implicit feedback.
\newblock In {\em Proceedings of the Twenty-Fifth Conference on Uncertainty in
  Artificial Intelligence}, pages 452--461. AUAI Press.

\bibitem[\protect\citename{Riedel \bgroup et al.\egroup }2013]{limin}
Sebastian Riedel, Limin Yao, Andrew McCallum, and Benjamin~M. Marlin.
\newblock 2013.
\newblock Relation extraction with matrix factorization and universal schemas.
\newblock In {\em HLT-NAACL}.

\bibitem[\protect\citename{Rocktaschel \bgroup et al.\egroup
  }2015]{logicmfnaacl15}
Tim Rocktaschel, Sameer Singh, and Sebastian Riedel.
\newblock 2015.
\newblock Injecting logical background knowledge into embeddings for relation
  extraction.
\newblock In {\em Annual Conference of the North American Chapter of the
  Association for Computational Linguistics (NAACL)}.

\bibitem[\protect\citename{Socher \bgroup et al.\egroup }2013]{socherkb}
Richard Socher, Danqi Chen, Christopher~D Manning, and Andrew Ng.
\newblock 2013.
\newblock Reasoning with neural tensor networks for knowledge base completion.
\newblock In {\em Advances in Neural Information Processing Systems.}

\bibitem[\protect\citename{Suchanek \bgroup et al.\egroup }2007]{yago}
Fabian~M. Suchanek, Gjergji Kasneci, and Gerhard Weikum.
\newblock 2007.
\newblock Yago: A core of semantic knowledge.
\newblock In {\em Proceedings of the 16th International Conference on World
  Wide Web}.

\bibitem[\protect\citename{Sukhbaatar \bgroup et al.\egroup
  }2015]{sukhbaatar2015end}
Sainbayar Sukhbaatar, Jason Weston, Rob Fergus, et~al.
\newblock 2015.
\newblock End-to-end memory networks.
\newblock In {\em Advances in Neural Information Processing Systems}, pages
  2431--2439.

\bibitem[\protect\citename{Toutanova \bgroup et al.\egroup
  }2015]{toutanova2015representing}
Kristina Toutanova, Danqi Chen, Patrick Pantel, Hoifung Poon, Pallavi
  Choudhury, and Michael Gamon.
\newblock 2015.
\newblock Representing text for joint embedding of text and knowledge bases.
\newblock In {\em Empirical Methods in Natural Language Processing (EMNLP)}.

\bibitem[\protect\citename{Verga \bgroup et al.\egroup
  }2016]{verga2015multilingual}
Patrick Verga, David Belanger, Emma Strubell, Benjamin Roth, and Andrew
  McCallum.
\newblock 2016.
\newblock Multilingual relation extraction using compositional universal
  schema.
\newblock {\em Annual Conference of the North American Chapter of the
  Association for Computational Linguistics (NAACL)}.

\bibitem[\protect\citename{Wang \bgroup et al.\egroup }2014]{wang2014knowledge}
Zhen Wang, Jianwen Zhang, Jianlin Feng, and Zheng Chen.
\newblock 2014.
\newblock Knowledge graph embedding by translating on hyperplanes.
\newblock In {\em Proceedings of the Twenty-Eighth AAAI Conference on
  Artificial Intelligence}, pages 1112--1119. Citeseer.

\bibitem[\protect\citename{Yang \bgroup et al.\egroup }2015]{bishan}
Bishan Yang, Wen{-}tau Yih, Xiaodong He, Jianfeng Gao, and Li~Deng.
\newblock 2015.
\newblock Embedding entities and relations for learning and inference in
  knowledge bases.
\newblock {\em International Conference on Learning Representations 2014}.

\bibitem[\protect\citename{Yao \bgroup et al.\egroup }2013]{yao2013universal}
Limin Yao, Sebastian Riedel, and Andrew McCallum.
\newblock 2013.
\newblock Universal schema for entity type prediction.
\newblock In {\em Proceedings of the 2013 workshop on Automated knowledge base
  construction}, pages 79--84. ACM.

\end{thebibliography}
